\documentclass[sigconf]{acmart}

\AtBeginDocument{%
  }

\copyrightyear{2026}
\acmYear{2026}
\setcopyright{cc}
\setcctype{by}
\acmConference[MM '26]{Proceedings of the 34th ACM International Conference on Multimedia}{November 10--14, 2026}{Rio de Janeiro, Brazil}
\acmBooktitle{Proceedings of the 34th ACM International Conference on Multimedia (MM '26), November 10--14, 2026, Rio de Janeiro, Brazil}
\acmDOI{10.1145/3767308.3835693}
\acmISBN{979-8-4007-2213-4/2026/11}
\usepackage{algorithmic}
\usepackage{algorithm}
\usepackage{graphicx}
\usepackage{textcomp}
\usepackage{booktabs}
\usepackage{multirow}
\usepackage{subcaption}
\usepackage{xcolor}
\usepackage{bm}
\usepackage{makecell}
\usepackage{balance}
\usepackage{float}
\usepackage{url}

\newcommand{\ours}{RIFT}
\newcommand{\oursfull}{Representation Inconsistency Forensics on Trajectories}
\newcommand{\RR}{\mathbb{R}}
\newcommand{\EE}{\mathbb{E}}
\newcommand{\zmacro}{\bm{z}^{\text{macro}}}
\newcommand{\zmicro}{\bm{z}^{\text{micro}}}

\title{Mind the Rift: Cross-Scale Coupling Mismatch \\ for AI-Generated Video Detection}
\titlenote{Accepted at the 34th ACM International Conference on Multimedia (MM~'26), November 10--14, 2026, Rio de Janeiro, Brazil. Published under CC~BY~4.0; Version of Record: \url{https://doi.org/10.1145/3767308.3835693}.}

\author{Siyu Li}
\orcid{0000-0003-2537-5055}
\affiliation{%
  \department{School of Cyber Science and Engineering}
  \institution{Sichuan University}
  \city{Chengdu}
  \state{Sichuan}
  \country{China}}
\email{lisiyu\_real@stu.scu.edu.cn}

\author{Jin Yang}
\authornote{Corresponding author.}
\orcid{0000-0001-8357-6571}
\affiliation{%
  \department{School of Cyber Science and Engineering}
  \institution{Sichuan University}
  \city{Chengdu}
  \state{Sichuan}
  \country{China}}
\affiliation{%
  \department{School of Information Science and Technology}
  \institution{Xizang University}
  \city{Lhasa}
  \state{Xizang}
  \country{China}}
\email{yangjin66@scu.edu.cn}

\author{Weiheng Liang}
\orcid{0009-0004-2257-298X}
\affiliation{%
  \department{School of Cyber Science and Engineering}
  \institution{Sichuan University}
  \city{Chengdu}
  \state{Sichuan}
  \country{China}}
\email{liangweiheng@stu.scu.edu.cn}

\begin{document}

\makeatletter
\@namedef{r@TotPages}{{9}{1}{}{}{}}
\makeatother

\begin{abstract}
As AI video generators achieve cinematic realism, reliable detection becomes essential for safeguarding digital trust. We identify cross-scale coupling mismatch as a new forensic signal, where scale refers to the level of abstraction (semantic dynamics vs.\ pixel-level residuals): in natural videos, macro-level temporal dynamics and micro-level residual patterns are intrinsically coupled by the unified imaging physics pipeline, whereas AI generators, whose training objectives do not explicitly preserve this joint distribution, systematically violate this coupling. Detecting such mismatch is challenging because it requires independently extracting information at both scales while simultaneously quantifying their cross-scale relationship. We propose \ours{} (\oursfull{}), an orthogonal forensic framework that addresses this through three interlocking components: a macro stream that builds a dynamic baseline of expected temporal evolution via differential geometry and persistent homology on learned manifold trajectories, a micro stream that acts as a sensitive forensic probe via steganalytic filtering and temporal modeling, and a coupling divergence module that measures the conditional dependency between the two streams. Gram-Schmidt orthogonality guarantees the information-theoretic validity of this measurement. Experiments on two benchmarks (VidProM, 120K videos, 7~generators; GenVidBench, 68K videos, 4~generators) demonstrate that \ours{} achieves 99.33\% and 99.72\% F1-score respectively, with 97.87\% unseen-generator detection rate in leave-one-out evaluation, while exhibiting encoder agnosticism: scaling from ViT-S/14 (22M) to ViT-L/14 (300M) changes F1 by less than 0.1\%, and switching to a different encoder family (DINOv1) reduces F1 by only 0.73\,pp. Code is available at \url{https://github.com/Litsay/RIFT}.
\end{abstract}

\ccsdesc[500]{Computing methodologies~Computer vision}
\ccsdesc[300]{Security and privacy~Digital rights management}
\ccsdesc[300]{Computing methodologies~Neural networks}

\keywords{AI-generated video detection, cross-scale coupling mismatch, video forensics, topological data analysis}

\maketitle

\renewcommand{\shortauthors}{Siyu Li, Jin Yang, and Weiheng Liang}

\section{Introduction}
\label{sec:intro}

Recent video generation models such as Sora~\cite{brooks2024sora}, Pika, and CogVideo~\cite{hong2023cogvideo} have reached a fidelity level where macro-level semantic errors and temporal inconsistencies are no longer prominent.
This progress poses a serious threat to digital trust: synthetic videos can circulate as authentic evidence, erode public confidence in visual media, and enable large-scale misinformation.

Existing detection methods have pursued three largely independent paradigms. \emph{Spatial detectors}~\cite{wang2020cnndetection,ojha2023universalfakedetect,wang2023dire,tan2024rethinking} identify artifacts in individual frames but ignore temporal structure. \emph{Temporal consistency detectors}~\cite{xu2023tall,ma2025decof,chen2024demamba} model frame-to-frame patterns but operate at a single semantic scale. \emph{Physics-informed detectors}~\cite{zhang2025nsgvd,interno2025restrv} leverage physical priors such as trajectory curvature in pretrained representation spaces, yet extract handcrafted statistics from one scale only. A recent line of work~\cite{he2026mpfnet} combines macro and micro analysis through a hierarchical dual-path architecture; however, the two branches operate as sequential, independent filters with no interaction, no orthogonality guarantee, and no mechanism to measure cross-scale coupling. \textbf{All these approaches share a common blind spot: they analyze signals at individual scales, missing the \emph{relationship} between scales as a forensic signal in its own right.}

\begin{figure}[t]
  \centering
  \includegraphics[width=0.94\linewidth]{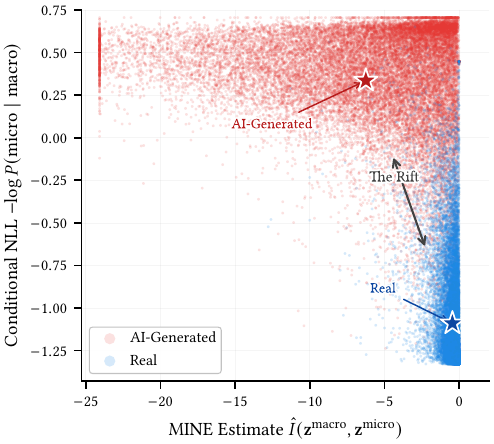}
  \caption{Empirical evidence of cross-scale coupling mismatch. Each point represents one video, plotted by its MINE-estimated mutual information $\hat{I}(\zmacro, \zmicro)$ (x-axis) and conditional negative log-likelihood $-\!\log P(\text{micro}\mid\text{macro})$ (y-axis), measured by a trained \ours{} model. Real videos (blue) cluster in the high-MI, low-NLL region (tightly coupled macro dynamics and micro residuals), whereas AI-generated videos (red) scatter toward low MI and high NLL (broken coupling). The gap between the two class centroids ($\star$) constitutes the ``rift'' that \ours{} exploits as a forensic signal (Cohen's $d = 4.46$ for NLL, $p < 10^{-300}$; $n = 120$K videos from VidProM and DVSC2023). The separation persists across encoder families and unseen generators (Secs.~\ref{sec:further}, \ref{sec:casestudy}): a robust, generator-agnostic regularity.}
  \label{fig:the_rift}
\end{figure}

We make a key observation: in natural videos, macro-level temporal dynamics (trajectory geometry in representation space) and micro-level residual patterns (pixel-domain noise fingerprints) are not independent but \emph{intrinsically coupled} by the unified imaging physics pipeline that produces both simultaneously. A camera's sensor noise correlates with its motion response; compression artifacts co-vary with scene complexity. AI generators, by contrast, optimize for perceptual quality and temporal coherence at individual scales without explicitly preserving their cross-scale coupling. Because the joint distribution of macro dynamics and micro residuals is never directly supervised during training, this coupling is systematically broken. We term this phenomenon \emph{cross-scale coupling mismatch} and characterize it empirically as a robust, generator-agnostic regularity: it manifests as a measurable divergence in the conditional dependency between macro and micro signals (Figure~\ref{fig:the_rift}) and persists across encoder families, training distributions, and all evaluated generators (Secs.~\ref{sec:further} and~\ref{sec:casestudy}).

To detect this mismatch, we propose \ours{} (\oursfull{}), an orthogonal forensic framework comprising three interlocking components. First, \emph{orthogonal decoupling} projects the shared representation into macro and micro subspaces with Gram-Schmidt constraints, ensuring independent information extraction. The \emph{macro stream} then builds a dynamic baseline of expected temporal evolution: it learns a low-dimensional manifold embedding~\cite{bengio2013representation,arvanitidis2018latent} on which differential geometry (curvature, torsion, jerk) and persistent homology characterize the expected dynamic patterns, building on the insight that temporal trajectories in learned representation spaces carry discriminative geometric structure~\cite{henaff2019perceptual}. The \emph{micro stream} serves as a sensitive forensic probe, extracting noise fingerprints via steganalytic filters and modeling their temporal dynamics. Finally, a \emph{coupling divergence module} measures $P(\text{micro} \mid \text{macro})$ through conditional likelihood estimation and mutual information, quantifying how far the observed macro-micro relationship deviates from the natural coupling baseline. Crucially, both streams are always active and contribute in parallel, unlike cascade approaches where one branch may be bypassed entirely. While the macro stream provides the structural backbone of detection, the coupling divergence between streams is the distinguishing mechanism: it accounts for the majority of \ours{}'s improvement over prior trajectory-based methods, separating the framework from single-scale approaches.

\noindent\textbf{Contributions.} Our main contributions are:
\begin{itemize}
    \item We identify \emph{cross-scale coupling mismatch} as a forensic signal: natural videos exhibit intrinsic coupling between macro-level temporal dynamics and micro-level residual patterns governed by imaging physics, which AI generators systematically violate.
    \item We propose \ours{} (\oursfull{}), an orthogonal forensic framework: the macro stream builds a dynamic baseline via manifold geometry and persistent homology, the micro stream serves as a forensic probe, and conditional dependency analysis measures coupling divergence with Gram-Schmidt orthogonality guarantees.
    \item We demonstrate encoder agnosticism: within the DINOv2 family, $14{\times}$ parameter scaling changes F1 by ${<}0.1\%$; across families (DINOv1 vs.\ DINOv2), F1 drops by only 0.73\,pp, proving that detection stems from framework design rather than encoder specifics.
    \item Experiments on VidProM~\cite{wang2024vidprom} (7~generators) and GenVidBench~\cite{ni2026genvidbench} (4~generators) show \ours{} achieves 99.33\% and 99.72\% F1 respectively, with 97.87\% unseen-generator detection in leave-one-out evaluation. The entire framework trains on a single consumer GPU (RTX~3080, 10\,GB).
\end{itemize}

\section{Method}
\label{sec:method}

\begin{figure*}[t]
  \centering
  \includegraphics[width=0.9\textwidth]{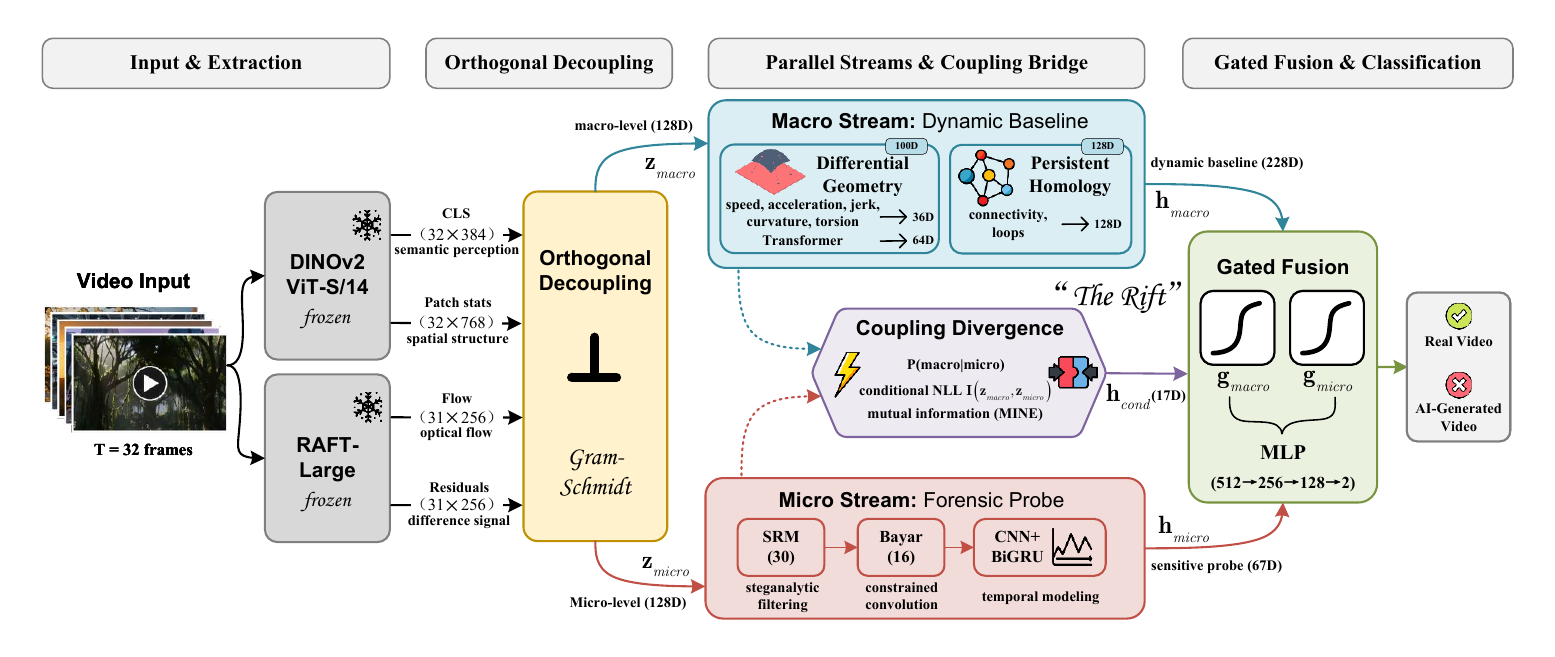}
  \caption{Overview of the \ours{} framework. Given a video, frozen encoders extract frame-level features, which are orthogonally decoupled into macro and micro subspaces. The macro stream (dynamic baseline) and micro stream (forensic probe) are analyzed independently, and their coupling divergence $P(\text{micro}\mid\text{macro})$ serves as the core forensic signal for classification.}
  \label{fig:architecture}
\end{figure*}

Given a video $V$ consisting of $T$ frames, our goal is to classify it as either authentic or AI-generated. The \ours{} framework proceeds in five stages (Figure~\ref{fig:architecture}): (1)~frozen backbone encoders extract multi-modal frame representations; (2)~orthogonal decoupling projects these into independent macro and micro subspaces; (3)~the macro stream builds a dynamic baseline characterizing expected temporal evolution; (4)~the micro stream extracts fine-grained forensic signals as a sensitive probe; and (5)~a coupling divergence module measures $P(\text{micro}\mid\text{macro})$ to quantify cross-scale mismatch. These signals are fused via learned gates for final classification.

\subsection{Feature Extraction}
\label{sec:feature_extraction}

We employ two frozen pretrained encoders. \textbf{DINOv2 ViT-S/14}~\cite{oquab2024dinov2} processes each frame at $224{\times}224$ resolution, producing CLS tokens $\bm{c}_t \in \RR^{384}$ capturing global semantics and patch-level statistics $\bm{p}_t \in \RR^{768}$ summarizing local spatial structure. \textbf{RAFT-Large}~\cite{teed2020raft} estimates dense optical flow between consecutive frames, yielding encoded flow vectors $\bm{f}_t \in \RR^{256}$ and motion-compensated residual vectors $\bm{r}_t \in \RR^{256}$. Both encoders remain completely frozen throughout training, ensuring that detection performance reflects downstream architectural design rather than encoder adaptation.

\subsection{Orthogonal Decoupling}
\label{sec:decoupling}

The coupling divergence measurement (Sec.~\ref{sec:conditional}) requires that macro and micro information be extracted \emph{independently}; otherwise, shared information would inflate divergence estimates with self-correlation rather than genuine cross-scale coupling. To enforce this, we project the concatenated frame representations $[\bm{c}_t; \bm{f}_t]$ and $[\bm{c}_t; \bm{r}_t; \bm{p}_t]$ through two MLP projectors $g_\text{macro}$ and $g_\text{micro}$, producing $\zmacro_t, \zmicro_t \in \RR^{d_z}$ ($d_z{=}128$).

Orthogonality is enforced through dual constraints. A \emph{soft} loss penalizes the average cosine similarity across the batch:
\begin{equation}
  \mathcal{L}_\text{ortho} = \frac{1}{BT}\sum_{b,t} \cos^2(\zmacro_{b,t},\, \zmicro_{b,t}).
  \label{eq:ortho}
\end{equation}
A \emph{hard} constraint applies Gram-Schmidt orthogonalization to the projector weight matrices every 100 training steps, preventing gradual drift.

To ensure that orthogonality does not discard information, a decoder reconstructs the original CLS token from the concatenation $[\zmacro_t; \zmicro_t]$:
\begin{equation}
  \mathcal{L}_\text{recon} = \frac{1}{BT}\sum_{b,t} \|\bm{c}_{b,t} - h_\text{dec}([\zmacro_{b,t}; \zmicro_{b,t}])\|^2.
  \label{eq:recon}
\end{equation}
The combination of orthogonality and reconstruction ensures that macro and micro subspaces partition the CLS representation space non-redundantly, providing a necessary condition for the coupling divergence to reflect genuine cross-scale dependency rather than self-correlation. Reconstruction targets the CLS representation (flow and patch features contribute directly to downstream losses), and the orthogonality guarantee applies to the coupling divergence measurement (Sec.~\ref{sec:conditional}); the micro stream's temporal features (Sec.~\ref{sec:micro}) are extracted independently from pixel-level residuals.

\subsection{Macro Stream: Dynamic Baseline}
\label{sec:macro}

The macro stream characterizes \emph{what normal temporal evolution looks like} in representation space. It builds a dynamic baseline from two complementary perspectives on the same trajectory.

\paragraph{Manifold Embedding.}
Computing differential geometry directly in $\RR^{d_z}$ is ill-conditioned: cosine similarity variance scales as $1/d$, so at $d_z{=}128$ direction changes collapse to near-constant values regardless of actual dynamics ($\operatorname{Var}[\cos\theta] \approx 0.008$). We address this through a learned embedding $\phi: \RR^{d_z} \to \RR^{d_m}$ (two-layer MLP, GELU):
\begin{equation}
  \bm{x}_t = \phi(\zmacro_t), \quad \bm{x}_t \in \RR^{d_m},\; d_m{=}24.
  \label{eq:manifold}
\end{equation}
At $d_m{=}24$, angular variance increases to ${\approx}0.042$, restoring discriminative geometric quantities. Beyond resolving angle concentration, $\phi$ learns a coordinate system adapted to temporal dynamics, consistent with the manifold hypothesis~\cite{bengio2013representation}, and absorbs encoder-specific distributional shifts, contributing to the encoder-agnostic property (Sec.~\ref{sec:further}).

\paragraph{Manifold Regularization.}
To ensure $\phi$ learns a faithful geometric structure, we regularize it through three complementary objectives. An \emph{isometry} term preserves local distance ratios during projection, where $\lambda_\text{iso}$ is a learnable scalar (initialized to~1):
\begin{equation}
  \mathcal{L}_\text{iso} = \frac{1}{T\!-\!1}\sum_{t} \biggl(\frac{\|\bm{x}_{t+1} - \bm{x}_t\|}{\|\zmacro_{t+1} - \zmacro_t\| + \epsilon} - \lambda_\text{iso}\biggr)^{\!2},
  \label{eq:iso}
\end{equation}
A \emph{smoothness} term, applied exclusively to real samples, penalizes discrete second-order differences to encode the inductive bias that authentic temporal evolution is smooth on the manifold:
\begin{equation}
  \mathcal{L}_\text{smooth} = \frac{1}{|\mathcal{B}_\text{real}|}\!\sum_{b \in \mathcal{B}_\text{real}} \frac{1}{T\!-\!2}\sum_{t} \|\bm{x}_{t+1} - 2\bm{x}_t + \bm{x}_{t-1}\|^2.
  \label{eq:smooth}
\end{equation}
An InfoNCE contrastive term~\cite{oord2018infonce} encourages temporally coherent geometry by clustering adjacent frames ($\operatorname{sim}$: cosine similarity, $\tau{=}0.15$):
\begin{equation}
  \mathcal{L}_\text{nce} = -\frac{1}{T\!-\!1}\sum_{t}\log\frac{\exp\bigl(\operatorname{sim}(\bm{x}_t, \bm{x}_{t+1})/\tau\bigr)}{\sum_{k \neq t}\exp\bigl(\operatorname{sim}(\bm{x}_t, \bm{x}_k)/\tau\bigr)},
  \label{eq:nce}
\end{equation}
The complete manifold loss is $\mathcal{L}_\text{manifold} = \mathcal{L}_\text{iso} + \mathcal{L}_\text{smooth} + \mathcal{L}_\text{nce}$.

\paragraph{Local: Differential Geometry.}
From the embedded trajectory $\bm{x}_1, \dots, \bm{x}_{T-1}$, we compute six groups of differential geometry quantities. Let $\Delta\bm{x}_t = \bm{x}_{t+1} - \bm{x}_t$ denote the discrete velocity. The \emph{Menger curvature}~\cite{leger1999menger} at each interior point quantifies local bending via the circumradius of three consecutive points:
\begin{equation}
  \kappa_t = \frac{2\sqrt{\|\Delta\bm{x}_{t\text{-}1}\|^2\|\Delta\bm{x}_t\|^2 - (\Delta\bm{x}_{t\text{-}1} \!\cdot\! \Delta\bm{x}_t)^2}}{\|\Delta\bm{x}_{t\text{-}1}\|\;\|\Delta\bm{x}_t\|\;\|\bm{x}_{t+1} - \bm{x}_{t-1}\|},
  \label{eq:curvature}
\end{equation}
where the numerator equals twice the triangle area of $(\bm{x}_{t-1}, \bm{x}_t, \bm{x}_{t+1})$; this definition is valid in any ambient dimension~$d_m$.
The \emph{discrete torsion} measures twisting out of the local osculating plane $\Pi_t = \operatorname{span}(\Delta\bm{x}_{t-1}, \Delta\bm{x}_t)$:
\begin{equation}
  \tau_t = \frac{\bigl\|\Delta\bm{x}_{t+1} - \operatorname{proj}_{\Pi_t}\!\bigl(\Delta\bm{x}_{t+1}\bigr)\bigr\|}{\|\Delta\bm{x}_t\|},
  \label{eq:torsion}
\end{equation}
capturing the component of the next velocity that is orthogonal to the current osculating plane. The \emph{direction change} captures the angular deviation between successive velocities:
\begin{equation}
  \delta_t = 1 - \frac{\Delta\bm{x}_t \cdot \Delta\bm{x}_{t+1}}{\|\Delta\bm{x}_t\|\;\|\Delta\bm{x}_{t+1}\|}.
  \label{eq:direction}
\end{equation}
This quantity captures the angular regularity of temporal evolution: the distribution of $\delta_t$ values differs systematically between natural and AI-generated videos, providing a discriminative geometric signal~\cite{henaff2019perceptual}. Together with \emph{speed} $\|\Delta\bm{x}_t\|$, \emph{acceleration} $\|\Delta^2\bm{x}_t\|$, and \emph{jerk} $\|\Delta^3\bm{x}_t\|$, these six quantities capture the complete local kinematic profile of the trajectory. Each group is summarized by six statistics (mean, std, max, min, kurtosis, spike ratio), producing a 36-dimensional geometry descriptor. A Transformer encoder (4~heads, 2~layers) models temporal patterns over the full curvature sequence, producing a 64-dimensional temporal embedding. The local perspective yields $\bm{h}_\text{local} \in \RR^{100}$.

\paragraph{Global: Topological Invariants.}
The same 24-dimensional trajectory is analyzed via persistent homology~\cite{edelsbrunner2010computational}.
We build a Vietoris--Rips filtration and compute persistence diagrams for $H_0$ (connected components) and $H_1$ (loops).
These diagrams are vectorized into persistence landscapes~\cite{bubenik2015statistical} with 4~functions and 16~bins per dimension ($\bm{h}_\text{global}\!\in\!\RR^{128}$).
Whereas differential geometry captures \emph{how} the trajectory bends locally, topology captures \emph{what} it is globally: a spiral and a figure-eight may share similar curvature yet differ in~$H_1$~structure.
These invariants are stable under continuous deformations.

The complete dynamic baseline is the concatenation $\bm{h}_\text{macro} = [\bm{h}_\text{local}; \bm{h}_\text{global}] \in \RR^{228}$.

\subsection{Micro Stream: Sensitive Forensic Probe}
\label{sec:micro}

The micro stream operates independently from the macro stream, extracting fine-grained forensic signals from pixel-level residuals that are invisible in semantic representations.

\paragraph{Noise Residual Extraction.}
For each frame $I_t$, we apply two complementary high-pass filter banks to suppress scene content while preserving generation artifacts. Steganalytic rich model (SRM) filters~\cite{fridrich2012rich} provide 30 predefined kernels $\{\bm{K}_i\}_{i=1}^{30}$ that extract linear and non-linear noise residuals. Bayar constrained convolutions~\cite{bayar2018constrained} complement these with 16 learned kernels $\bm{W}_j$ whose weights satisfy:
\begin{equation}
  \bm{W}_j(0,0) = -\!\!\!\!\sum_{(i,k) \neq (0,0)}\!\! \bm{W}_j(i,k),
  \label{eq:bayar}
\end{equation}
so that each kernel acts as a prediction error filter: the center pixel is predicted from its spatial neighbors, and only the prediction residual passes through. The constraint is re-enforced after each gradient step by renormalizing the center weight. The combined per-frame forensic feature is $\bm{u}_t = [\bm{r}_t^{\text{srm}};\, \bm{r}_t^{\text{bayar}};\, \bm{s}_t]$, where $\bm{r}_t^{\text{srm}}$ and $\bm{r}_t^{\text{bayar}}$ are pooled residual statistics from the respective filter banks and $\bm{s}_t$ aggregates spatial and frequency-domain statistics of the motion-compensated residual $\bm{r}_t$ from RAFT (Sec.~\ref{sec:feature_extraction}).

\paragraph{Temporal Modeling.}
A convolutional encoder reduces the dimensionality of each $\bm{u}_t$, and a bidirectional GRU models the temporal dynamics of the resulting sequence, yielding a 64-dimensional temporal embedding $\bm{h}_\text{temp}$ from the concatenated final hidden states. We append three scalar temporal consistency metrics computed from the frame-to-frame forensic changes $d_t = \|\bm{u}_{t+1} - \bm{u}_t\|$:
\begin{equation}
  \mu_d = \tfrac{1}{T\!-\!1}\!\sum_{t} d_t,\,
  \sigma_d^2 = \tfrac{1}{T\!-\!1}\!\sum_{t} (d_t - \mu_d)^2,\,
  \rho_d = \frac{\sum_{t} (d_t\!-\!\mu_d)(d_{t+1}\!-\!\mu_d)}{\sum_{t} (d_t - \mu_d)^2}.
  \label{eq:consistency}
\end{equation}
These scalars capture the magnitude, regularity, and serial dependence of forensic signal fluctuations, respectively. The micro stream produces $\bm{h}_\text{micro} = [\bm{h}_\text{temp};\, \mu_d;\, \sigma_d^2;\, \rho_d] \in \RR^{67}$.

\subsection{Coupling Divergence Measurement}
\label{sec:conditional}

This module is the \emph{distinguishing mechanism} of \ours{}: it measures the cross-scale coupling between the macro baseline and the micro probe. The hypothesis is that in natural videos, the macro temporal state constrains the distribution of micro observations through shared imaging physics; in AI-generated videos, this constraint is violated.

\paragraph{Conditional Prediction.}
An MLP predicts the expected distribution of micro statistics given the macro state:
\begin{equation}
  \mu_t, \log\sigma^2_t = f_\text{cond}(\zmacro_t), \quad
  \text{NLL}_t = \tfrac{1}{2}\bigl(\log\sigma^2_t + (\bm{s}_t - \mu_t)^2 / \sigma^2_t\bigr),
  \label{eq:cond_nll}
\end{equation}
where $\bm{s}_t$ denotes the micro statistics sequence from the micro stream. The per-video NLL statistics (mean, max, std) are projected to a 16-dimensional feature vector, encoding how well the macro state predicts micro behavior: low NLL indicates strong coupling, while high NLL indicates mismatch.

\paragraph{Mutual Information Estimation.}
A MINE network~\cite{belghazi2018mine} estimates the mutual information $I(\zmacro; \zmicro)$ by contrasting the joint distribution with a shuffled marginal:
\begin{equation}
  \hat{I} = \EE_\text{joint}[T_\theta(\zmacro, \zmicro)] - \log\EE_\text{marginal}[e^{T_\theta(\zmacro, \tilde{\zmicro})}],
  \label{eq:mine}
\end{equation}
where $\tilde{\zmicro}$ is obtained by temporal shuffling. This scalar quantifies the residual statistical dependency between the orthogonal subspaces.

\paragraph{Why Orthogonality Matters.}
Without the orthogonal guarantee from Sec.~\ref{sec:decoupling}, the coupling divergence degenerates: shared information between $\zmacro$ and $\zmicro$ would produce artificially high MI and artificially low NLL regardless of the input's authenticity. Orthogonality ensures that any measured coupling reflects \emph{genuine cross-scale dependency}, which is precisely the signal that AI generators fail to replicate.

The module outputs $\bm{h}_\text{cond} \in \RR^{17}$ (16-dimensional projected NLL features concatenated with the 1-dimensional MI estimate).

\subsection{Gated Fusion and Classification}
\label{sec:fusion}

Two projection heads map the streams to a common 256D space.
The macro path projects $\bm{h}_\text{macro} \in \RR^{228}$ and the micro path projects $[\bm{h}_\text{micro}; \bm{h}_\text{cond}] \in \RR^{84}$.
Learned sigmoid gates $g_\text{macro}, g_\text{micro} \in (0,1)$ weight the two paths:
\begin{equation}
  \bm{h}_\text{fused} = g_\text{macro} \cdot \bm{h}'_\text{macro} + g_\text{micro} \cdot \bm{h}'_\text{micro}.
\end{equation}
The gates are independent (not softmax-normalized), allowing the model to upweight both streams simultaneously when both provide reliable signals. A 3-layer MLP classifier ($512 \to 256 \to 128 \to 2$) produces the final binary prediction.

\subsection{Training Strategy}
\label{sec:training}

Training proceeds in two stages to stabilize the orthogonal decomposition before joint optimization.

\paragraph{Stage~1: Decoupling Pretraining (30 epochs).} Only the orthogonal decoupling module and its associated manifold regularization are trained, with learning rate $10^{-3}$. This establishes well-separated macro and micro subspaces before downstream modules are introduced.

\paragraph{Stage~2: Joint Training (50 epochs).} All trainable modules are optimized jointly with learning rate $3{\times}10^{-4}$, cosine annealing with warm restarts ($T_0{=}5$, $T_\text{mult}{=}2$), and 5\% linear warmup. The total loss combines five components:
\begin{equation}
  \mathcal{L} = \mathcal{L}_\text{focal} + \alpha\,\mathcal{L}_\text{ortho} + \beta\,\mathcal{L}_\text{recon} + \gamma\,\mathcal{L}_\text{manifold} + \delta\,\mathcal{L}_\text{cond},
  \label{eq:total_loss}
\end{equation}
where $\mathcal{L}_\text{focal}$ is the focal loss~\cite{lin2017focal} ($\gamma_\text{focal}{=}2.0$), $\mathcal{L}_\text{manifold}$ is the manifold regularization defined in Sec.~\ref{sec:macro} (Eqs.~\ref{eq:iso}--\ref{eq:nce}), and $\mathcal{L}_\text{cond}$ is the conditional NLL applied only to real samples to learn the natural coupling distribution. Weights: $\alpha{=}0.1$, $\beta{=}0.5$, $\gamma{=}0.02$, $\delta{=}0.05$. We use AdamW~\cite{loshchilov2019adamw} with weight decay $0.01$ and gradient clipping at norm $1.0$. Backbone parameters remain frozen; only 4.7M trainable parameters are updated.

\section{Experiments}
\label{sec:exp}

\subsection{Experimental Setup}
\label{sec:setup}

\paragraph{Dataset.}
We construct an evaluation set of 120K videos from VidProM~\cite{wang2024vidprom} and DVSC2023~\cite{pizzi2023dvsc}, the same public sources used by prior work~\cite{interno2025restrv,he2026mpfnet}. The \emph{real} partition consists of 50K videos from DVSC2023, capturing diverse real-world content. The \emph{fake} partition contains 70K videos generated by seven text-to-video models from VidProM: Pika~\cite{pika2024}, VideoCrafter2~\cite{chen2024videocrafter2}, Text2Video-Zero~\cite{khachatryan2023text2video}, ModelScope~\cite{wang2023modelscope}, CogVideo~\cite{hong2023cogvideo}, OpenSora~\cite{zheng2024opensora}, and Show-1 (${\sim}$10K each). Prior VidProM evaluations typically include 4~generators; we expand coverage to all 7 available generators and adopt a 50/50 stratified train/test split with a fixed seed, holding out 10\% of the training partition for validation. All baselines in Table~\ref{tab:main} are re-evaluated under this same protocol; image-level detectors are aggregated to video-level predictions by majority voting over frame-level decisions.

We additionally evaluate on GenVidBench~\cite{ni2026genvidbench}, an independent benchmark of 68K videos spanning 4~generators (CogVideo, Mora, MuseV, SVD~\cite{blattmann2023svd}) with Main (general scenes) and Plants (high-randomness) subsets, using the same 50/50 protocol.

\paragraph{Implementation.}
All features are precomputed and cached: DINOv2 ViT-S/14 CLS tokens $(32, 384)$, patch statistics $(32, 768)$, RAFT-encoded flow $(31, 256)$, and motion-compensated residuals $(31, 256)$ per video, totaling ${\sim}20$\,GB. Topology features are precomputed via Vietoris-Rips persistent homology and cached separately. Training uses a single NVIDIA RTX~3080 (10\,GB). The two-stage training procedure, all hyperparameters, and loss weights are detailed in Appendix~\ref{app:hyperparams}. Source code, pretrained models, and Seedance~2.0 demo videos are available at \url{https://github.com/Litsay/RIFT}.

\paragraph{Metrics.}
We report Accuracy, Precision, Recall, F1-score, and AUC. F1 is the primary metric for model selection.

\subsection{Main Results and Cross-Generator Generalization}
\label{sec:main_results}

\paragraph{Comparison with Prior Methods.}
Table~\ref{tab:main} compares \ours{} against 10~baselines from four paradigms, all reproduced under identical conditions (Sec.~\ref{sec:setup}).

Image-level detectors~\cite{wang2020cnndetection,ojha2023universalfakedetect,doloriel2024fredect,liu2020gramnet} remain near chance level on both benchmarks (F1 $\leq$ 67\%), confirming that per-frame analysis alone cannot capture the temporal structure essential for video-level detection. Temporal detectors~\cite{xu2023tall,gu2021stil,chen2024demamba} improve substantially but plateau below 93\% F1. The strongest prior results come from physics-informed methods (ReStraV~\cite{interno2025restrv}: 97.83\% F1 on VidProM, 95.88\% F1 on GenVidBench) and multi-branch approaches (MPF-Net~\cite{he2026mpfnet}: 97.03\% F1 on GenVidBench).

\ours{} achieves state-of-the-art results on both benchmarks: 99.33\% F1 on VidProM with 7~generators ($+$1.50\,pp over ReStraV) and 99.72\% F1 on GenVidBench ($+$2.69\,pp over MPF-Net). Per-generator detection rates range from 100\% (T2VZero, VideoCrafter2, ModelScope) to 97.24\% (CogVideo) on VidProM, with full breakdowns in Appendix~\ref{app:per_generator}.

\begin{table}[t]
\centering
\caption{Comparison with prior methods on VidProM~\cite{wang2024vidprom} and GenVidBench~\cite{ni2026genvidbench}. All methods are reproduced from open-source implementations under identical conditions on both benchmarks (7~generators, 50/50 stratified split on VidProM; 50/50 split on GenVidBench).}
\label{tab:main}
\small
\setlength{\tabcolsep}{2.5pt}
\begin{tabular}{llcccccc}
\toprule
 & \multirow{2}{*}{Method} & \multicolumn{3}{c}{VidProM} & \multicolumn{3}{c}{GenVidBench} \\
\cmidrule(lr){3-5} \cmidrule(lr){6-8}
 & & Acc & F1 & AUC & Acc & F1 & AUC \\
\midrule
\multicolumn{8}{l}{\emph{Image-level detectors}} \\
& CNNDetection~\cite{wang2020cnndetection} & 50.41 & 53.82 & 52.63 & 55.81 & 53.31 & 57.94 \\
& UnivFD~\cite{ojha2023universalfakedetect} & 51.27 & 55.80 & 57.62 & 58.56 & 60.04 & 60.92 \\
& FreDect~\cite{doloriel2024fredect} & 60.98 & 59.73 & 63.56 & 59.36 & 60.73 & 61.25 \\
& Gram-Net~\cite{liu2020gramnet} & 63.64 & 66.75 & 64.22 & 66.59 & 65.50 & 68.22 \\
\midrule
\multicolumn{8}{l}{\emph{Temporal video detectors}} \\
& TALL~\cite{xu2023tall} & 73.89 & 75.66 & 77.31 & 77.36 & 78.58 & 81.06 \\
& STIL~\cite{gu2021stil} & 75.42 & 75.28 & 76.96 & 74.92 & 73.43 & 76.33 \\
& DeMamba~\cite{chen2024demamba} & 93.58 & 92.74 & 95.58 & 92.78 & 91.37 & 94.64 \\
\midrule
\multicolumn{8}{l}{\emph{Physics-informed \& trajectory-based}} \\
& ReStraV~\cite{interno2025restrv} & 97.06 & 97.83 & 98.39 & 96.93 & 95.88 & 98.24 \\
& NSG-VD~\cite{zhang2025nsgvd} & 95.37 & 96.41 & 96.37 & 93.27 & 94.98 & 94.63 \\
\midrule
\multicolumn{8}{l}{\emph{Multi-branch}} \\
& MPF-Net~\cite{he2026mpfnet} & 95.13 & 94.37 & 96.39 & 96.35 & 97.03 & 97.34 \\
\midrule
& \textbf{\ours{} (Ours)} & \textbf{99.22} & \textbf{99.33} & \textbf{99.96} & \textbf{99.56} & \textbf{99.72} & \textbf{99.96} \\
\bottomrule
\end{tabular}
\end{table}

\paragraph{Cross-Generator Generalization.}
To evaluate generalization to unseen generators, we conduct Leave-One-Generator-Out (LOGO) experiments across all 7~generators: one generator is entirely excluded from training, and the model is evaluated on all its test samples. Table~\ref{tab:logo} shows that \ours{} maintains an average unseen-generator detection rate of 97.87\%. Four generators (T2VZero, VideoCrafter2, ModelScope, Show-1) are detected at ${\geq}$99.4\%, while the hardest cases (OpenSora 94.91\%, CogVideo 95.47\%, Pika 95.50\%) still exceed 94\%.
\begin{table}[t]
\centering
\caption{Leave-One-Generator-Out generalization on VidProM across all 7~generators. ``Unseen'' denotes the detection rate on the excluded generator's test samples.}
\label{tab:logo}
\small
\begin{tabular}{@{}lccc@{}}
\toprule
Excluded Generator & Unseen Det.\ Rate & Seen Test F1 & Seen Test AUC \\
\midrule
T2VZero        & 100.00\% & 98.77\% & 99.88\% \\
VideoCrafter2  & 100.00\% & 98.91\% & 99.89\% \\
ModelScope     & 99.78\%  & 98.86\% & 99.89\% \\
Show-1         & 99.40\%  & 98.84\% & 99.89\% \\
Pika           & 95.50\%  & 98.95\% & 99.92\% \\
CogVideo       & 95.47\%  & 99.54\% & 99.96\% \\
OpenSora       & 94.91\%  & 98.95\% & 99.91\% \\
\midrule
\textbf{Average} & \textbf{97.87\%} & 98.97\% & 99.91\% \\
\bottomrule
\end{tabular}
\end{table}

\subsection{Ablation Study}
\label{sec:ablation}

We ablate each conceptual component of \ours{} on VidProM by zeroing its features at the fusion input while retraining Stage~2 from a shared Stage~1 checkpoint. Table~\ref{tab:ablation} reports results on the held-out test set.

\begin{table}[t]
\centering
\caption{Stream-level ablation on VidProM (50/50 test set, ${\sim}$60K samples). ``Macro only'' retains both local geometry and global topology; ``Micro only'' retains only the micro stream.}
\label{tab:ablation}
\small
\begin{tabular}{lcccc}
\toprule
Configuration & Acc & F1 & AUC & $\Delta$F1 \\
\midrule
Full \ours{} & 99.22 & 99.33 & 99.96 & --- \\
\midrule
w/o Geometry & 97.78 & 97.96 & 98.89 & $-$1.37 \\
w/o Coupling & 97.85 & 98.01 & 98.90 & $-$1.32 \\
\midrule
Macro only & 97.82 & 97.99 & 98.88 & $-$1.34 \\
Micro only & 80.89 & 83.54 & 89.50 & $-$15.79 \\
\bottomrule
\end{tabular}
\end{table}

Three findings stand out. \textbf{(1)~The macro stream is the structural backbone}: the macro stream alone (Macro only) achieves 97.99\% F1, while the micro stream in isolation reaches only 83.54\% F1 ($-$15.79\,pp), confirming that manifold geometry and topology carry the majority of discriminative information. Notably, w/o Geometry (97.96\%) performs slightly below Macro only (97.99\%) despite retaining the micro stream and coupling module, indicating that coupling depends on a strong macro baseline: without local geometry, the macro state is too weak for reliable conditional prediction. \textbf{(2)~Coupling divergence is the key complementary signal}: removing only the divergence module reduces F1 by 1.32\,pp, nearly as large as the gap between Full and Macro only ($-$1.34\,pp). This means the micro stream's independent contribution beyond coupling is only 0.02\,pp; virtually all of its value flows through the conditional dependency $P(\text{micro}\mid\text{macro})$, directly validating the coupling mismatch hypothesis. Moreover, since Macro only (97.99\%) barely exceeds the strongest prior method (ReStraV, 97.83\%), the coupling mechanism accounts for 88\% of \ours{}'s total improvement over the state of the art ($+$1.32 of $+$1.50\,pp), confirming its role as the distinguishing component of the framework.
Within the macro stream, adding topological invariants to local differential geometry further improves F1, confirming that persistent homology captures complementary global structure.

\subsection{Encoder Agnosticism and Further Analysis}
\label{sec:further}

\paragraph{Encoder Ablation.}
A key property of \ours{} is encoder agnosticism: detection remains strong regardless of encoder capacity or architecture family (Table~\ref{tab:encoder}). Within the DINOv2 family, scaling from ViT-S/14 (22M) to ViT-L/14 (300M) changes F1 by only 0.02\%. Across families, replacing DINOv2 ViT-S/14 with DINOv1~\cite{caron2021dino} ViT-S/16 (a different pretraining method, patch size, and training data) reduces F1 by just 0.73\,pp (99.33\% $\to$ 98.60\%) while AUC drops only 0.12\,pp (99.96\% $\to$ 99.84\%). This demonstrates that the coupling mismatch signal is captured by the framework's orthogonal decomposition and geometric analysis, not by any specific encoder's representation power. The manifold embedding $\phi$ (Sec.~\ref{sec:macro}) absorbs encoder-specific distributional shifts, enabling consistent geometric analysis across architecturally diverse backbones.

\begin{table}[t]
\centering
\caption{Encoder agnosticism on VidProM. Within the DINOv2 family, scaling from 22M to 300M parameters changes F1 by only 0.02\%. Across families (DINOv1 vs.\ DINOv2), F1 drops by just 0.73\,pp.}
\label{tab:encoder}
\small
\begin{tabular}{llcccc}
\toprule
Family & Encoder & Params & Acc & F1 & AUC \\
\midrule
\multirow{2}{*}{DINOv2} & ViT-S/14 & 22M & 99.22\% & 99.33\% & 99.96\% \\
 & ViT-L/14 & 300M & 99.28\% & 99.35\% & 99.98\% \\
\midrule
DINOv1 & ViT-S/16 & 21M & 98.37\% & 98.60\% & 99.84\% \\
\bottomrule
\end{tabular}
\end{table}

\paragraph{Handcrafted Feature Integration.}
To test whether handcrafted geometry features provide complementary information, we append the 21-dimensional curvature-and-distance statistics from~\cite{interno2025restrv} to the macro stream input. F1 drops by 0.63\,pp, indicating that \ours{}'s learned manifold geometry strictly subsumes these handcrafted descriptors, and their addition introduces redundancy into the fusion layer.

\paragraph{Efficiency.}
Table~\ref{tab:latency} reports module-level latency on a single RTX 3080 GPU. The macro stream's Transformer encoder dominates at 86\% of post-backbone time. Including backbone extraction, end-to-end latency is ${\sim}$182\,ms (ViT-S/14) or ${\sim}$461\,ms (ViT-L/14), with 4.7M trainable parameters.

\begin{table}[t]
\centering
\caption{Module-level latency profiling (precomputed features).}
\label{tab:latency}
\small
\begin{tabular}{lrc}
\toprule
Module & Latency (ms) & \% Total \\
\midrule
Orthogonal Decoupling & 1.6 & 3.4 \\
Macro Stream (Transformer) & 40.4 & 86.0 \\
Micro Stream (SRM+GRU) & 1.9 & 4.1 \\
Coupling Divergence & 2.2 & 4.6 \\
Gated Fusion & 0.9 & 2.0 \\
\midrule
\textbf{Post-backbone Total} & \textbf{47.0} & 100.0 \\
\bottomrule
\end{tabular}
\end{table}

\subsection{Interpretability}
\label{sec:interpret}

\begin{figure}[t]
  \centering
  \includegraphics[width=0.88\linewidth]{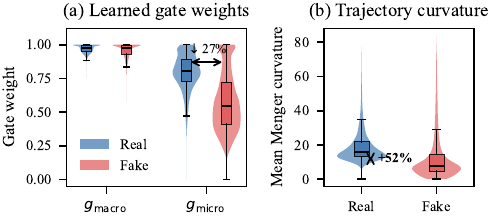}
  \caption{Interpretability of \ours{}. (a)~Learned gate weights:
  $g_\text{macro}$ remains near-saturated for both classes, while
  $g_\text{micro}$ drops 27\% for AI-generated videos, showing
  autonomous suppression of the less reliable probe.
  (b)~Mean Menger curvature: real trajectories exhibit 52\% higher
  curvature on the learned manifold, indicating that natural videos
  produce richer geometric dynamics while AI-generated videos
  exhibit overly uniform trajectories.}
  \label{fig:interpret}
\end{figure}

\paragraph{Coupling Mismatch Validation.}
To directly validate the coupling mismatch hypothesis, we collect per-sample mutual information $\hat{I}(\zmacro, \zmicro)$ and conditional NLL from the trained model across all 120K videos. Real videos exhibit significantly higher MI ($-0.45 \pm 0.68$) than AI-generated videos ($-6.25 \pm 5.95$), with Cohen's $d = 1.37$ and $p < 10^{-300}$ (Mann-Whitney U). (Negative MI estimates reflect MINE's loose variational bound at low true MI, but relative class separation remains valid.) The conditional NLL shows an even stronger separation: real $-1.09 \pm 0.30$ versus fake $+0.34 \pm 0.34$ (Cohen's $d = 4.46$). The orthogonality check confirms that $\cos(\zmacro, \zmicro) \approx 0.001$ for both classes, ruling out information leakage. This separation is consistent across all 7 generators (see Appendix~\ref{app:coupling}), indicating that coupling mismatch is a robust, generator-agnostic regularity of current AI video generation rather than an artifact of specific architectures. Figure~\ref{fig:the_rift} visualizes this result.

\paragraph{Gate Weight Analysis.}
The macro gate remains near-saturated for both classes ($g_\text{macro} \approx 0.96$; Figure~\ref{fig:interpret}a), while the micro gate drops from $0.80$ (real) to $0.58$ (fake), a 27\% relative suppression. This emergent asymmetry provides independent evidence for the coupling mismatch hypothesis: the model autonomously downweights the micro probe when cross-scale coupling is broken.

\paragraph{Trajectory Geometry.}
The Menger curvature of learned manifold trajectories shows a clear distributional shift (Figure~\ref{fig:interpret}b): real videos exhibit 52\% higher mean curvature than AI-generated videos ($18.99 \pm 8.12$ vs.\ $12.49 \pm 6.73$). This indicates that natural videos produce richer, more structured temporal dynamics on the learned manifold, while AI-generated videos exhibit overly uniform trajectories characteristic of diffusion-based synthesis. Yet the substantial distributional overlap confirms that coupling divergence, not individual geometry, provides the decisive signal.

\subsection{Robustness Analysis}
\label{sec:robustness}

We evaluate \ours{} on VidProM under 17 degradation conditions simulating real-world post-processing: H.264 and H.265 re-encoding (500\,kbps to 8\,Mbps), Gaussian noise ($\sigma=5{-}20$), and spatial cropping (70\%--90\%). Features are re-extracted from degraded videos and evaluated with the frozen model.

\ours{} achieves peak robustness under H.265 re-encoding at 2\,Mbps (F1 = 88.19\%) and tolerates Gaussian noise well at most levels (F1 $>$ 88\% at $\sigma \in \{5, 15, 20\}$), though performance varies non-monotonically across degradation intensities on the 996-sample evaluation subset. Performance degrades under aggressive spatial cropping (F1\,$<$\,0.77 at 80\% crop), because cropping alters the spatial context on which patch statistics and optical flow are computed. Full per-generator results are provided in Appendix~\ref{app:robustness}.

\subsection{Case Study: Zero-Shot Detection of Seedance 2.0}
\label{sec:casestudy}

To evaluate whether the coupling mismatch hypothesis generalizes beyond the training distribution, we conduct a zero-shot case study on Seedance 2.0~\cite{bytedance2026seedance}, a commercial video generation model released by ByteDance in February 2026 that was entirely absent from our training data. We collect 27 videos spanning diverse content categories and evaluate them with the frozen \ours{} model without any fine-tuning or adaptation.

\begin{figure}[t]
  \centering
  \includegraphics[width=0.88\linewidth]{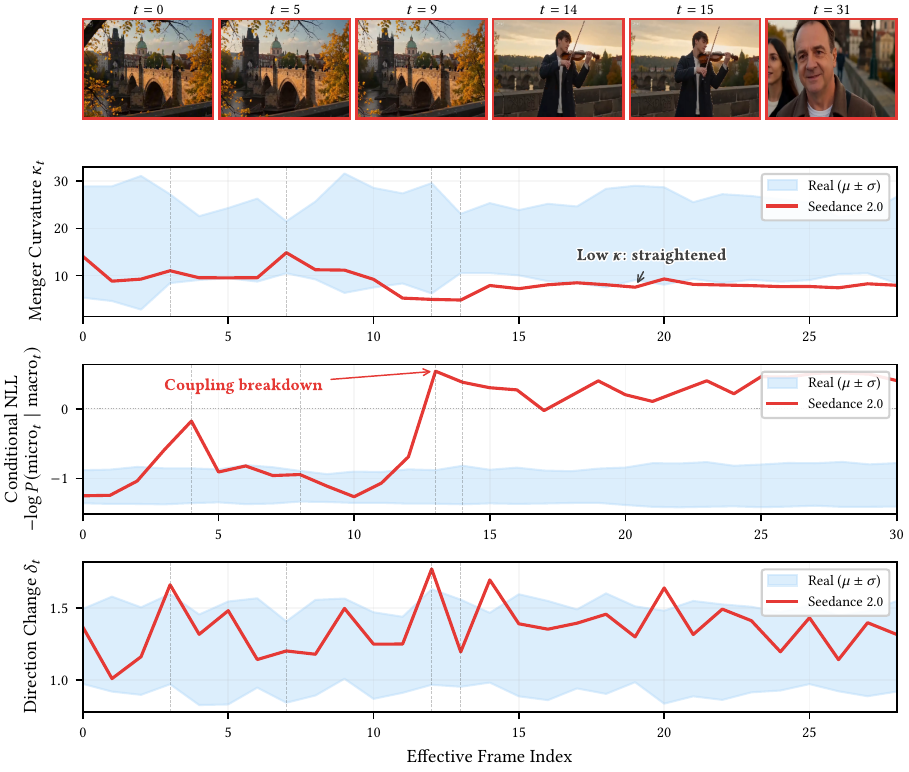}
  \caption{Per-frame analysis of a Seedance 2.0 video (street musician). Top: sampled frames. Bottom three rows: Menger curvature $\kappa_t$, conditional NLL, and direction change $\delta_t$, each compared against the real-video baseline ($\mu \pm \sigma$, blue). Seedance 2.0 exhibits lower curvature (overly uniform dynamics), elevated NLL (coupling breakdown), and uniform $\delta_t$ (unnatural regularity).}
  \label{fig:deepdive}
\end{figure}

\paragraph{Detection and Coupling Signals.}
\ours{} correctly classifies all 27 videos (100\% detection rate; binomial 95\% CI [86.8\%, 100\%], $n{=}27$; mean confidence 97.5\%, range 73.4\%--100.0\%) despite zero exposure to this generator during training. Comparing the coupling signals against the seven training generators, Seedance 2.0 exhibits even weaker cross-scale coupling in MI ($\hat{I} = {-}17.74$ vs.\ fake avg $-6.25$) and micro gate suppression ($g_\text{micro} = 0.49$ vs.\ $0.58$), while its conditional NLL ($+0.30$) falls within the fake distribution ($+0.34 \pm 0.34$). The extreme MI value may partially reflect MINE estimator instability on out-of-distribution inputs and should be read as a relative indicator of coupling breakdown.

\paragraph{Per-Frame Analysis.}
Figure~\ref{fig:deepdive} presents per-frame analysis of a representative video against a real-video baseline ($\mu \pm \sigma$, 30 videos). The Menger curvature $\kappa_t$ lies consistently below the real band (mean $10.28$ vs.\ $18.99$), consistent with the overly uniform trajectory pattern observed across AI-generated videos. The conditional NLL reveals localized coupling breakdowns, and the direction change $\delta_t$ shows unnaturally uniform temporal dynamics characteristic of diffusion-based generators.

\section{Related Work}
\label{sec:related}

\subsection{AI-Generated Image and Video Detection}

Detection of AI-generated visual content has evolved through three largely independent paradigms. \emph{Spatial detectors}~\cite{wang2020cnndetection,ojha2023universalfakedetect,wang2023dire,tan2024rethinking} identify per-frame artifacts in frequency, feature, or reconstruction space but discard temporal structure. \emph{Temporal consistency detectors}, most recently DeMamba~\cite{chen2024demamba}, model frame-to-frame patterns but operate at a single semantic scale~\cite{xu2023tall,ma2025decof,gu2021stil,zhang2023ummaformer}. \emph{Physics-informed detectors} such as NSG-VD~\cite{zhang2025nsgvd} and ReStraV~\cite{interno2025restrv} leverage physical priors such as spatiotemporal gradients or trajectory curvature~\cite{henaff2019perceptual}, yet extract handcrafted statistics from one scale only. Orthogonally, recent efforts address diffusion-specific artifacts~\cite{cheng2024diffusion}, cross-domain generalization to unseen forgery methods~\cite{xia2024advancing}, grounded artifact reasoning~\cite{li2025skyra}, and training-free detection via second-order temporal features~\cite{zheng2025d3}. All these paradigms analyze signals at individual scales; \ours{} instead targets the cross-scale coupling relationship itself.

\subsection{Hierarchical and Multi-Branch Detection}

Recent work has begun to combine macro and micro analysis. Shuai et al.~\cite{shuai2023locate} propose a two-stream network that fuses localization and verification branches through multi-scale collaborative learning. MPF-Net~\cite{he2026mpfnet} introduces a hierarchical dual-path framework with a manifold sentinel branch and a micro-temporal fluctuation branch, while Zhou et al.~\cite{zhou2025vfmgun} demonstrate that frozen VFMs inherently possess off-manifold detection capabilities. Three fundamental distinctions separate these approaches from \ours{}: (1)~MPF-Net's branches operate \emph{sequentially and independently} with no cross-branch interaction, whereas \ours{} runs both streams \emph{in parallel with orthogonal constraints}; (2)~each branch answers a self-contained question, whereas \ours{} measures the coupling divergence $P(\text{micro}\mid\text{macro})$ as the core forensic evidence; and (3)~MPF-Net's performance varies across VFM backbones, whereas \ours{} exhibits encoder agnosticism (${<}0.1\%$ within DINOv2, $0.73$\,pp across encoder families).

\subsection{Topological Data Analysis in Temporal Modeling}

Persistent homology~\cite{edelsbrunner2010computational} provides algebraic invariants that characterize the shape of data across scales: connected components ($H_0$) capture fragmentation and merging, while loops ($H_1$) capture cyclic structures. Persistence landscapes~\cite{bubenik2015statistical} offer a stable, vectorized summary compatible with machine learning pipelines. While topological data analysis has been applied to time series classification and shape recognition, to our knowledge \ours{} is the first to apply persistent homology to AI-generated video detection. In our framework, Vietoris--Rips filtrations on learned manifold trajectories complement local differential geometry with global structural invariants (Sec.~\ref{sec:macro}).

\section{Conclusion}
\label{sec:conclusion}

We have presented \ours{}, an orthogonal forensic framework built on a central hypothesis: AI video generators, by optimizing perceptual quality at individual scales without explicit cross-scale supervision, systematically break the cross-scale coupling that the unified imaging physics pipeline imposes on natural videos. This coupling mismatch constitutes a structural limitation of current generative architectures, because the joint distribution across scales is never explicitly optimized during generation. Our framework detects this mismatch through orthogonal macro-micro decomposition, a geometric and topological dynamic baseline, an independent forensic probe, and the coupling divergence between the two streams.

Experimentally, \ours{} achieves 99.33\% F1 on VidProM (7~generators) and 99.72\% F1 on GenVidBench (4~generators), with 97.87\% unseen-generator detection rate in leave-one-out evaluation. A key property is encoder agnosticism: $14{\times}$ parameter scaling within DINOv2 changes F1 by ${<}0.1\%$, and switching to DINOv1 reduces F1 by only 0.73\,pp. With only 4.7M trainable parameters and frozen pretrained encoders, the entire pipeline trains on a single consumer GPU (RTX~3080, 10\,GB) and achieves ${\sim}$182\,ms end-to-end inference latency.

\paragraph{Limitations and Future Work.}
\ours{} exhibits fragility under aggressive spatial cropping, which alters the spatial context on which patch statistics and optical flow are computed; spatial augmentation during feature extraction (random cropping) is a direct remedy. The macro Transformer accounts for 86\% of post-backbone compute, motivating distillation into a compact student model. Finally, broader evaluation on next-generation architectures beyond the Seedance 2.0 case study (Sec.~\ref{sec:casestudy}), including Sora~\cite{brooks2024sora} and Kling, will further assess the generality of the coupling mismatch hypothesis.

\begin{acks}
This work was supported by the National Natural Science Foundation of China (Grants No.~62162057 and No.~61872254), by the Key Laboratory of Data Protection and Intelligent Management of the Ministry of Education, China (Grant No.~SCUSAKFKT202402Y), and by the Open Program of Shaanxi Key Laboratory of Intelligent Policing (Independent Research Topic, Grant No.~SXZJ25KF08).
\end{acks}

\bibliographystyle{ACM-Reference-Format}
\balance
\bibliography{references}


\appendix

\section{Training Details and Hyperparameters}
\label{app:hyperparams}

\subsection{Two-Stage Training Procedure}

\ours{} employs a two-stage training strategy with frozen backbone encoders (DINOv2 ViT-S/14 and RAFT-Large):

\paragraph{Stage 1: Orthogonal Decoupling Pretraining (30 epochs).}
Only the orthogonal decoupling module (macro/micro projectors and reconstruction decoder) and the macro stream analyzer are trainable; the fusion, conditional dependency, and micro stream modules are frozen. The manifold regularization weight is elevated ($\gamma_{\text{manifold}} = 0.2$) to encourage well-structured manifold trajectories before joint training. Learning rate: $10^{-3}$ with linear warmup (5\% of epochs) followed by cosine annealing. Gram-Schmidt hard orthogonalization is applied every 100 optimization steps.

\paragraph{Stage 2: Joint Training (50 epochs).}
All trainable modules are unfrozen (backbone encoders remain frozen). The manifold weight is reduced to $\gamma_{\text{manifold}} = 0.02$ to let classification loss dominate. Learning rate: $3 \times 10^{-4}$ with linear warmup followed by cosine annealing with warm restarts ($T_0 = 5$, $T_{\text{mult}} = 2$). Early stopping with patience of 20 epochs on validation F1-macro.

\subsection{Hyperparameter Settings}

Table~\ref{tab:hyperparams} summarizes all hyperparameters used in the final model.

\begin{table}[h]
\centering
\caption{Complete hyperparameter settings for \ours{}.}
\label{tab:hyperparams}
\footnotesize
\setlength{\tabcolsep}{2pt}
\begin{tabular}{@{}llc@{}}
\toprule
Category & Parameter & Value \\
\midrule
\multirow{7}{*}{Model} & Orthogonal embedding $d_z$ & 128 \\
 & Manifold dimension $d_{\text{manifold}}$ & 24 \\
 & Micro statistics $d_{\text{stat}}$ & 128 \\
 & Macro temporal $d_{\text{temporal}}$ & 64 \\
 & Micro temporal $d_{\text{temporal\_micro}}$ & 64 \\
 & Topology $d_{\text{topo}}$ (4 landscapes $\times$ 16 bins $\times$ 2 dims) & 128 \\
 & InfoNCE temperature $\tau$ & 0.15 \\
\midrule
\multirow{4}{*}{Sampling} & Segment length $L$ & 32 frames \\
 & Spatial resolution & $224 \times 224$ \\
 & Segments per video (train) & 1 \\
 & Segments per video (eval) & 3 \\
\midrule
\multirow{5}{*}{Loss Weights} & Orthogonality $\alpha_{\text{ortho}}$ & 0.1 \\
 & Reconstruction $\beta_{\text{recon}}$ & 0.5 \\
 & Manifold $\gamma_{\text{manifold}}$ (Stage 1 / Stage 2) & 0.2 / 0.02 \\
 & Conditional $\delta_{\text{cond}}$ & 0.05 \\
 & Focal loss $\gamma_{\text{focal}}$ & 2.0 \\
\midrule
\multirow{6}{*}{Training} & Stage 1 epochs / learning rate & 30 / $10^{-3}$ \\
 & Stage 2 epochs / learning rate & 50 / $3 \times 10^{-4}$ \\
 & Batch size & 64 \\
 & Weight decay & 0.01 \\
 & Gradient clipping & 1.0 \\
 & Gram-Schmidt interval & 100 steps \\
\midrule
\multirow{3}{*}{Regularization} & Fusion dropout & 0.15 \\
 & Conditional predictor dropout & 0.2 \\
 & Micro GRU dropout & 0.15 \\
\midrule
\multirow{3}{*}{Augmentation} & Gaussian blur prob. & 0.2 \\
 & Resize degradation prob. & 0.3 \\
 & Random crop prob. & 0.3 \\
\bottomrule
\end{tabular}
\end{table}

\subsection{Loss Function}

The total training loss is:
\begin{multline}
\mathcal{L} = \mathcal{L}_{\text{focal}} + \alpha \mathcal{L}_{\text{ortho}} + \beta \mathcal{L}_{\text{recon}} \\
+ \gamma (\mathcal{L}_{\text{iso}} + \mathcal{L}_{\text{smooth}} + \mathcal{L}_{\text{nce}}) + \delta \mathcal{L}_{\text{cond}}
\end{multline}

where $\mathcal{L}_{\text{focal}}$ is class-balanced focal loss ($\gamma_{\text{focal}}{=}2.0$, equal class weights), $\mathcal{L}_{\text{ortho}}$ penalizes cosine similarity between streams, $\mathcal{L}_{\text{recon}}$ reconstructs CLS tokens, $\mathcal{L}_{\text{iso}}$/$\mathcal{L}_{\text{smooth}}$/$\mathcal{L}_{\text{nce}}$ regularize the manifold (isometry, smoothness, contrastive), and $\mathcal{L}_{\text{cond}}$ is the conditional prediction loss (real samples only).

\section{Per-Generator Detection Rates}
\label{app:per_generator}

\subsection{VidProM}

Table~\ref{tab:per_gen_vidprom} reports per-generator detection rates on the VidProM test set (50/50 stratified split, ${\sim}$60K samples).

\begin{table}[h]
\centering
\caption{Per-generator detection on VidProM test set (50/50 stratified split). Per-generator rates use single-segment evaluation; aggregate metrics in Table~1 of the main text use the standard multi-segment protocol.}
\label{tab:per_gen_vidprom}
\small
\begin{tabular}{lrcc}
\toprule
Generator & $n$ & Detection Rate & False Negatives \\
\midrule
T2VZero & 4,953 & 100.00\% & 0 \\
VideoCrafter2 & 4,968 & 100.00\% & 0 \\
ModelScope & 5,008 & 100.00\% & 0 \\
Show-1 & 5,012 & 99.90\% & 5 \\
Pika & 5,019 & 99.08\% & 46 \\
OpenSora & 5,066 & 98.36\% & 83 \\
CogVideo & 4,970 & 97.24\% & 137 \\
\midrule
Real (FPR) & 25,000 & --- & 470 FP (1.88\%) \\
\midrule
\textbf{Overall} & \textbf{59,996} & \textbf{99.23\%} & \textbf{271 FN total} \\
\bottomrule
\end{tabular}
\end{table}

CogVideo and OpenSora are the most challenging generators (97.24\% and 98.36\% respectively), likely because their generation pipelines produce more temporally coherent outputs that better preserve cross-scale coupling patterns.

\subsection{GenVidBench}

Table~\ref{tab:per_gen_gvb} reports per-generator results on GenVidBench (50/50 split, ${\sim}$34K samples).

\begin{table}[h]
\centering
\caption{Per-generator detection on GenVidBench.}
\label{tab:per_gen_gvb}
\small
\begin{tabular}{llccc}
\toprule
Subset & Generator & Acc & F1 & Det.\ Rate \\
\midrule
\multirow{5}{*}{Main} & CogVideo & 99.61\% & 99.80\% & 99.61\% \\
 & Mora & 99.97\% & 99.99\% & 99.97\% \\
 & MuseV & 99.95\% & 99.98\% & 99.95\% \\
 & SVD & 99.97\% & 99.98\% & 99.97\% \\
 & Real & 98.27\% & --- & 1.73\% FPR \\
\midrule
\multirow{5}{*}{Plants} & CogVideo & 100.00\% & 100.00\% & 100.00\% \\
 & Mora & 100.00\% & 100.00\% & 100.00\% \\
 & MuseV & 100.00\% & 100.00\% & 100.00\% \\
 & SVD & 100.00\% & 100.00\% & 100.00\% \\
 & Real & 98.10\% & --- & 1.90\% FPR \\
\bottomrule
\end{tabular}
\end{table}

\ours{} achieves near-perfect detection on all GenVidBench generators, with CogVideo being the most challenging on the Main subset (99.61\%), consistent with VidProM findings. The Plants subset (high visual randomness) is perfectly detected at 100\% for all generators.

\section{Full Robustness Results}
\label{app:robustness}

Table~\ref{tab:robustness_full} presents \ours{}'s performance under all 17 degradation conditions evaluated on a 996-sample VidProM subset. Features are re-extracted from degraded videos and evaluated with the frozen model.

\begin{table}[h]
\centering
\caption{Full robustness evaluation under 17 degradation conditions. Clean baseline: F1 = 99.05\%.}
\label{tab:robustness_full}
\small
\begin{tabular}{llccc}
\toprule
Category & Condition & Acc & F1 & AUC \\
\midrule
\emph{Clean} & None & 98.90 & 99.05 & 99.89 \\
\midrule
\multirow{5}{*}{H.264} & 500 kbps & 62.95 & 75.68 & 55.31 \\
 & 1 Mbps & 70.08 & 67.82 & 93.00 \\
 & 2 Mbps & 80.62 & 85.43 & 95.00 \\
 & 4 Mbps & 74.40 & 81.59 & 85.24 \\
 & 8 Mbps & 64.36 & 76.32 & 65.66 \\
\midrule
\multirow{5}{*}{H.265} & 500 kbps & 63.86 & 75.77 & 61.04 \\
 & 1 Mbps & 75.28 & 81.53 & 86.87 \\
 & 2 Mbps & 87.18 & 88.19 & 95.32 \\
 & 4 Mbps & 77.60 & 81.19 & 86.98 \\
 & 8 Mbps & 74.97 & 83.76 & 82.30 \\
\midrule
\multirow{4}{*}{Gaussian noise} & $\sigma = 5$ & 88.45 & 90.09 & 93.95 \\
 & $\sigma = 10$ & 70.87 & 79.89 & 78.07 \\
 & $\sigma = 15$ & 88.57 & 90.51 & 94.35 \\
 & $\sigma = 20$ & 84.56 & 88.30 & 95.65 \\
\midrule
\multirow{3}{*}{Spatial crop} & 70\% & 59.64 & 35.34 & 85.42 \\
 & 80\% & 70.00 & 76.79 & 87.67 \\
 & 90\% & 60.86 & 68.88 & 80.02 \\
\bottomrule
\end{tabular}
\end{table}

\paragraph{Analysis.}
H.265 generally outperforms H.264 at matched bitrates, peaking at 2\,Mbps (F1 = 88.19\%). Gaussian noise is tolerated at most levels (F1 $> 88\%$ at $\sigma \in \{5, 15, 20\}$). Some conditions exhibit non-monotonic behavior (e.g., $\sigma{=}10$ drops to 79.89\%, H.264 1\,Mbps to 67.82\%), which we attribute to the limited evaluation subset ($n = 996$) and the interaction between specific degradation parameters and the precomputed feature pipeline. Spatial cropping is the most damaging degradation (F1 drops to 35.34\% at 70\% crop), because cropping alters the spatial context upon which patch statistics and optical flow are computed, disrupting the feature extraction pipeline rather than the classifier itself.

\section{Per-Generator Coupling Mismatch Statistics}
\label{app:coupling}

Table~\ref{tab:coupling_per_gen} reports per-generator coupling statistics computed from the trained model across all 119,991 training+test videos.

\begin{table}[h]
\centering
\caption{Per-generator coupling mismatch statistics. MI = MINE mutual information estimate; NLL = conditional negative log-likelihood; $\cos(\mathbf{z}_{\text{macro}}, \mathbf{z}_{\text{micro}})$ = orthogonality check; $\bar{\kappa}$ = mean Menger curvature. Curvature statistics in this table are computed over all 119,991 videos. The main text (Sec.~3.5) reports values from a 24K interpretability subset with balanced per-generator sampling, which yields slightly different statistics for both classes: real $18.99 \pm 8.12$ (vs.\ $19.04 \pm 9.43$ here) and fake $12.49 \pm 6.73$ (vs.\ $12.49 \pm 12.59$ here). The larger fake standard deviation in this table reflects cross-generator variance.}
\label{tab:coupling_per_gen}
\small
\setlength{\tabcolsep}{2.5pt}
\begin{tabular}{@{}lrcccr@{}}
\toprule
Source & $n$ & MI & NLL & $|\cos|$ & $\bar{\kappa}$ \\
\midrule
\textbf{Real} & 49,999 & $-0.45 \pm 0.68$ & $-1.09 \pm 0.30$ & 0.0009 & $19.04 \pm 9.43$ \\
\midrule
CogVideo & 10,000 & $-6.31 \pm 5.42$ & $+0.24 \pm 0.30$ & 0.0005 & $11.87 \pm 8.43$ \\
ModelScope & 10,000 & $-4.58 \pm 4.92$ & $+0.01 \pm 0.47$ & 0.0008 & $5.02 \pm 2.01$ \\
OpenSora & 10,000 & $-4.61 \pm 5.10$ & $+0.44 \pm 0.24$ & 0.0021 & $23.53 \pm 14.95$ \\
Pika & 9,993 & $-3.86 \pm 4.02$ & $+0.55 \pm 0.21$ & 0.0004 & $28.87 \pm 17.27$ \\
Show-1 & 10,000 & $-10.16 \pm 6.25$ & $+0.30 \pm 0.16$ & 0.0021 & $9.74 \pm 5.38$ \\
T2VZero & 9,999 & $-9.57 \pm 7.45$ & $+0.31 \pm 0.33$ & 0.0015 & $1.50 \pm 0.31$ \\
VideoCrafter2 & 10,000 & $-4.63 \pm 4.17$ & $+0.50 \pm 0.24$ & 0.0017 & $7.11 \pm 2.54$ \\
\midrule
\textbf{Fake avg} & 69,992 & $-6.25 \pm 5.95$ & $+0.34 \pm 0.34$ & 0.0013 & $12.49 \pm 12.59$ \\
\bottomrule
\end{tabular}
\end{table}

\paragraph{Key Observations.}
(1)~\emph{Universal coupling breakdown}: all 7 generators exhibit substantially lower MI and positive NLL compared to real videos, confirming that coupling mismatch is a universal property of current AI video generation.
(2)~\emph{Orthogonality is maintained}: $|\cos(\mathbf{z}_{\text{macro}}, \mathbf{z}_{\text{micro}})| < 0.003$ for all generators, confirming no information leakage between streams.
(3)~\emph{Generator-specific signatures}: curvature varies widely across generators (T2VZero: $1.50$, Pika: $28.87$), reflecting different temporal dynamics. Despite this diversity, the coupling divergence signal remains consistently discriminative.
(4)~\emph{NLL is the strongest separator}: Cohen's $d = 4.46$ for NLL vs.\ $1.37$ for MI, because NLL directly measures the conditional dependency $P(\text{micro} \mid \text{macro})$.

\section{Seedance 2.0 Full Results}
\label{app:seedance}

Table~\ref{tab:seedance_full} reports per-video detection results for all 27 Seedance 2.0 videos evaluated in zero-shot mode (no fine-tuning or adaptation).

\begin{table}[H]
\centering
\caption{Per-video Seedance 2.0 zero-shot detection results. All 27 videos correctly classified as AI-generated. Coupling statistics: MI = $-17.74$, NLL = $+0.30$, $g_{\text{micro}} = 0.49$, $\bar{\kappa} = 10.28$.}
\label{tab:seedance_full}
\small
\begin{tabular}{lcc}
\toprule
Video Content & Pred & Confidence \\
\midrule
Aerial terraces & Fake & 97.1\% \\
Ancient architecture & Fake & 99.6\% \\
Animals (beach) & Fake & 100.0\% \\
Animals (grassland) & Fake & 100.0\% \\
Aurora, Iceland & Fake & 99.8\% \\
Calligraphy & Fake & 100.0\% \\
Children, playground & Fake & 96.1\% \\
City traffic & Fake & 98.8\% \\
Coffee still life & Fake & 89.5\% \\
Desert landscape & Fake & 90.8\% \\
Fireworks night & Fake & 100.0\% \\
Football & Fake & 100.0\% \\
Indoor cooking & Fake & 100.0\% \\
Industrial factory & Fake & 100.0\% \\
Macro flowers & Fake & 98.8\% \\
Musician & Fake & 99.0\% \\
Nature coastline & Fake & 100.0\% \\
Ocean sailboat & Fake & 100.0\% \\
Pastoral rice fields & Fake & 100.0\% \\
Rain scene & Fake & 100.0\% \\
Snow forest & Fake & 99.9\% \\
Sports person & Fake & 100.0\% \\
Starry timelapse & Fake & 89.9\% \\
Street music & Fake & 99.8\% \\
Tech robot & Fake & 99.7\% \\
Underwater ocean & Fake & 73.4\% \\
Waterfall jungle & Fake & 100.0\% \\
\midrule
\textbf{Mean / All} & \textbf{27/27} & \textbf{97.5\%} \\
\bottomrule
\end{tabular}
\end{table}

The lowest-confidence video is \emph{underwater ocean} (73.4\%), featuring slow-moving marine life with minimal motion structure that weakens the macro baseline's geometric features. Nevertheless, all 27 videos are correctly classified, demonstrating that the coupling mismatch signal generalizes to unseen commercial generators.

\paragraph{Coupling Signal Comparison.}
Seedance 2.0 exhibits even weaker cross-scale coupling than the 7 training generators: MI = $-17.74$ (vs.\ fake avg $-6.25$), NLL = $+0.30$ (within fake distribution $+0.34 \pm 0.34$), $g_{\text{micro}} = 0.49$ (vs.\ fake avg $0.58$, real $0.80$), and $\bar{\kappa} = 10.28$ (vs.\ fake avg $12.49$, real $19.04$). The extreme MI value may partially reflect MINE estimator instability on out-of-distribution data; it should be interpreted as a relative indicator of coupling breakdown rather than a calibrated measurement.

\end{document}